\documentclass[journal]{IEEEtran}

\ifCLASSINFOpdf
\else
\fi

\usepackage{lineno,amsmath,amsfonts}
\usepackage{color}
\usepackage{multirow}
\usepackage{algorithm}
\usepackage{algorithmic}
\usepackage{subfigure}
\usepackage{graphicx}
\usepackage[numbers,sort&compress]{natbib}
\usepackage{soul}
\usepackage[pagebackref=true,breaklinks=true,letterpaper=true,colorlinks,bookmarks=false]{hyperref}

\newcommand{\etal}{\textit{et al}.}
\newcommand{\eg}{\textit{e.g}.}
\newcommand{\ie}{\textit{i.e}.}

\begin{document}

\title{Towards better Validity: Dispersion based Clustering for Unsupervised Person Re-identification}

\author{Guodong~Ding,
        Salman~Khan,  
        Zhenmin~Tang, 
        Jian~Zhang,~\IEEEmembership{Senior Member,~IEEE} and
        Fatih~Porikli,~\IEEEmembership{Fellow,~IEEE}
\thanks{Guodong Ding and Zhenmin Tang are with the School of Computer Science and Engineering, Nanjing University of Science and Technology, Nanjing, Jiangsu, China. e-mails: guodong.ding@njust.edu.cn and tzm.cs@njust.edu.cn.}
\thanks{Salman Khan is with the College of Engineering and Computer Science, Australian National University, Australia. e-mail: salman.khan@anu.edu.au}
\thanks{Jian Zhang is with the School of Electrical and Data Engineering, University of Technology Sydney, Australia. e-mail: jian.zhang@uts.edu.au}
\thanks{Fatih Porikli is with the Research School of Engineering, Australian National University, Australia. e-mail: fatih.porikli@anu.edu.au}}

\markboth{Journal of \LaTeX\ Class Files,~Vol.~xx, No.~x, August~2019}%
{Shell \MakeLowercase{\textit{et al.}}: Bare Demo of IEEEtran.cls for IEEE Journals}

\maketitle

\begin{abstract}

Person re-identification aims to establish the correct identity correspondences of a person moving through a non-overlapping multi-camera installation. Recent advances based on deep learning models for this task mainly focus on supervised learning scenarios where accurate annotations are assumed to be available for each setup. Annotating large scale datasets for person re-identification is demanding and burdensome, which renders the deployment of such supervised approaches to real-world applications infeasible. Therefore, it is necessary to train models without explicit supervision in an autonomous manner. 

In this paper, we propose an elegant and practical clustering approach for unsupervised person re-identification based on the cluster validity consideration. Concretely, we explore a fundamental concept in statistics, namely \emph{dispersion}, to achieve a robust clustering criterion. Dispersion reflects the compactness of a cluster when employed at the intra-cluster level and reveals the separation when measured at the inter-cluster level. With this insight, we design a novel Dispersion-based Clustering (DBC) approach which can discover the underlying patterns in data. This approach considers a wider context of sample-level pairwise relationships to achieve a robust cluster affinity assessment which handles the complications may arise due to prevalent imbalanced data distributions. Additionally, our solution can automatically prioritize standalone data points and prevents inferior clustering. 
Our extensive experimental analysis on image and video re-identification benchmarks demonstrate that our method outperforms the state-of-the-art unsupervised methods by a significant margin. Code is available at \url{https://github.com/gddingcs/Dispersion-based-Clustering.git}.

\end{abstract}


\section{Introduction}
\label{sec:intro}

The goal of the person re-identification (re-ID) is to find the correspondences of the same person across a system of non-overlapping cameras. It has critical applications such as people tracking and mobility analysis in multi-camera streams. To this end, plenty of research work in recent years investigated supervised learning schemes, mainly leveraging on neural networks models, which led to remarkable performance gains \cite{yang2018person,cho2018pamm,feng2018learning,suh2018part,sun2017svdnet,zhao2017spindle,zhou2017point}. However, supervised learning requires large annotated datasets with manually labeled identities, which is a laborious undertaking for complex scenes. This motivated unsupervised approaches that are often based on handcrafted features \cite{farenzena2010person,liao2015person,lisanti2015person}, saliency indicators \cite{zhao2013unsupervised,wang2014unsupervised} and sparsity constraints \cite{kodirov2015dictionary}. These attempts, on the other hand, attain inferior performance compared to fully supervised models.

As an alternative, a variety of methods treats the person re-ID as an unsupervised domain adaptation task \cite{fan2018unsupervised,peng2016unsupervised,deng2018image}. The main idea is first to learn an identity embedding using an auxiliary dataset where the ground truth labels are available, and later transfer these learned feature representations onto an unlabeled target dataset. However, this strategy relies on the premise that these two domains share the same identity label space. For person re-ID, this assumption does not always hold since the identities from two different datasets are usually disjoint. 

Clustering based techniques have also been studied within the context of person re-ID. For instance, Fan \etal\ \cite{fan2018unsupervised} proposed pre-training a convolutional neural network (CNN) model on an external re-ID dataset where they apply $k$-means clustering on the features extracted from a target dataset to progressively select highly reliable data points for updating the model. One shortcoming of this approach is that the number of clusters, which dictates to the number of people, is preset and its right choice is unknown in the runtime. 

\begin{figure}[tb]
\begin{center}
    \centering
    \includegraphics[width=0.48\textwidth]{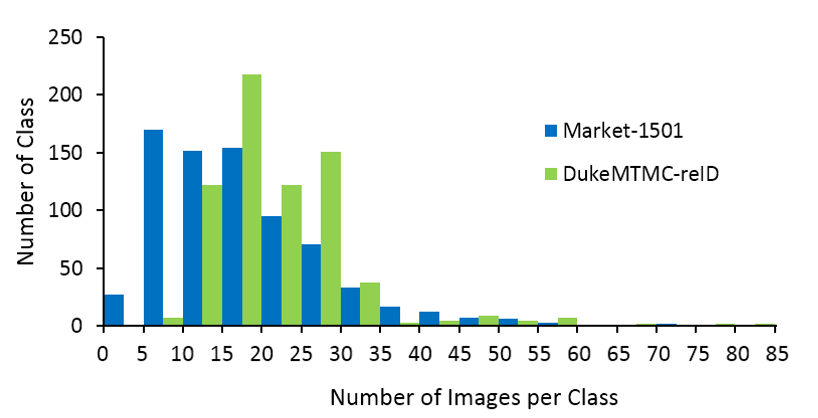}
    \caption{The number of classes vs. the number of images per class for Market-1501 and DukeMTMC-reID datasets. The distributions are imbalanced, which is common for person re-ID scenarios. }
    \label{fig:distribution}
    \end{center}
\end{figure}

Person re-ID methods based on clustering schemes often have a hierarchical nature, and some can also avoid the use of external datasets. As an example, the inspiring work in Lin \etal\ \cite{lin2019bottom} presented a bottom-up clustering (BUC) approach that alternatively trains a CNN model and performs merging without any dependence on auxiliary data samples. It uses the minimum distance between images in two clusters as the similarity measure for the merging operation. However, such a naive heuristic is suboptimal since it only considers one pair of images from two clusters discarding other potentially useful cues. This naive scheme may merge distinct identity groups by forming elongated clusters, which result in poor performance. Besides, it uses a diversity regularization term, which is defined as the number of samples in a cluster, to impose isometric clusters. In reality, data distribution over classes (identities) is usually imbalanced with a long tail; thus such a regularization term can be error-prone and invalid. To illustrate this issue, we show the number of samples per class on two re-ID datasets Market-1501 and DukeMTMC-reID in Fig.~\ref{fig:distribution}. As visible, the isometric cluster presumption does not hold for person re-ID scenarios. 

We tackle this challenging problem by incorporating a cluster validity criterion. Cluster validity is a measure to assess the quality of clustering results. We consider a clustering to be valid if it satisfies two dispersion based properties. In statistics, dispersion is the extent to which a distribution is stretched or compressed; thus, it characterizes the clustering quality. Low intra-dispersion and high inter-dispersion are the indicators of valid and questionable clusters. 

Motivated by this, we employ this elegant and straightforward criterion as our merging rule. It consists of two balancing objectives: an inter-cluster dispersion term and an intra-cluster dispersion term. The inter-cluster term quantifies the average linkage between the clusters while measuring their separation in a designated feature space; hence, the clusters with low dispersion are placed higher in the merging list. We empirically show that the average linkage models a broader context, which makes it superior to single linkage strategy used in \cite{lin2019bottom}. The intra-cluster term evaluates the compactness of candidate clusters and serves as a regularizer complementing the former term. Different from \cite{lin2019bottom}, which considers a cluster size constraint, we note that the number of samples should not be the primary concern as long as they are close enough to be considered as the same identity. As such, the intra-dispersion also helps to bypass the potential class imbalance issue.

In addition to introducing the above dispersion based criterion, we employ an agglomerative and alternating learning procedure tailored for the person re-ID task. The images of people moving in a non-overlapping multi-camera system typically constitute compact clusters according to their identities and acquisition conditions for each camera. However, viewpoint variations cause significant intra-class variations alongside. To address this challenge, we exploit a feature learning framework that trains CNN models and performs clustering in an alternating fashion. For CNN model training, we utilize the repelled loss that can incorporate both inter-class and intra-class variances to obtain invariant feature representations. With one accord, our clustering reduces intra-cluster dispersion while maximizing inter-cluster dispersion. The resulting clusters attain more refined structures, which further enhances the representation capability of the CNN model. Therefore, CNN model training and clustering leverage each other in a reciprocal exchange. 

The cluster dispersion term brings two significant advantages to our approach. It allows automatically prioritizing singletons and preventing inadequate clustering. Singleton refers to a cluster with only one sample. Ideally, in a multi-camera system designed to re-identify people, a singleton cluster should not exist. Our approach selectively assigns a higher priority to a singleton cluster when the candidate clusters have equal inter-dispersion since a singleton has no intra-dispersion. Our criterion also minimizes potentially incorrect decisions of the agglomerative merging steps by deferring high intra-dispersion clusters form merging.

Our contributions are three-fold:
\begin{itemize}
    \item  We propose to use cluster validity as the guidance and derive a dispersion based criterion which promotes compact and well-separated clusters.
    \item The proposed criterion automatically handles singleton cluster problem and can prevent poor clusters. Furthermore, our approach has a faster learning speed and is more stable than its counterpart BUC \cite{lin2019bottom}.
    \item The experimental results demonstrate that our approach outperforms the state-of-the-art unsupervised methods on both image-based and video-based re-ID datasets by a significant margin.
\end{itemize}

The remainder of this paper is organized as follows. In Section \ref{sec:related}, we review related work. In Section \ref{sec:approach}, we provide details of our proposed unsupervised approach with cluster dispersion followed by a discussion on why DBC works better. Section \ref{sec:experiments} demonstrates the effectiveness of proposed approach on both image-based and video-based person re-ID datasets along with an extensive ablation study. We conclude our work in Section \ref{sec:conclusion}.

\section{Related Work}
\label{sec:related}

Top-performing deep architectures are trained on massive amounts of labeled data. Most existing re-ID models are trained with human annotated ID labels in a supervised mode. Therefore, their deployment in real-world applications is usually hindered by lack of large-scale annotated training sets. To learn models without explicit supervision has therefore been extensively studied in the literature.

\subsection{Traditional Unsupervised Solutions}

 Some unsupervised methods with hand-crafted features have been proposed in recent years \cite{farenzena2010person,khan2016unsupervised,kodirov2016person,kodirov2015dictionary,lisanti2015person,liu2017stepwise,ma2017person,wang2014unsupervised,wang2016towards,ye2019dynamic,zhao2017person,yao2019deep}. However, they achieve inadequate re-ID performance when compared to the supervised learning methods. Specifically, Farenzena \etal\ \cite{farenzena2010person} exploited the property of symmetry in person images to deal with view variances. To handle the illumination changes and cluttered background, Ma \etal\ \cite{ma2012bicov} proposed to combine the Gabor filters and the covariance descriptor. Fisher Vector is explored in \cite{ma2012local} to encode higher order statistics of local features. Kodvirov \etal\ \cite{kodirov2015dictionary} proposed to combine a laplacian regularization term with conventional dictionary learning formulation to encode cross-view correspondences.

\begin{figure*}[htb]
\begin{center}
    \centering
    \includegraphics[width=\textwidth]{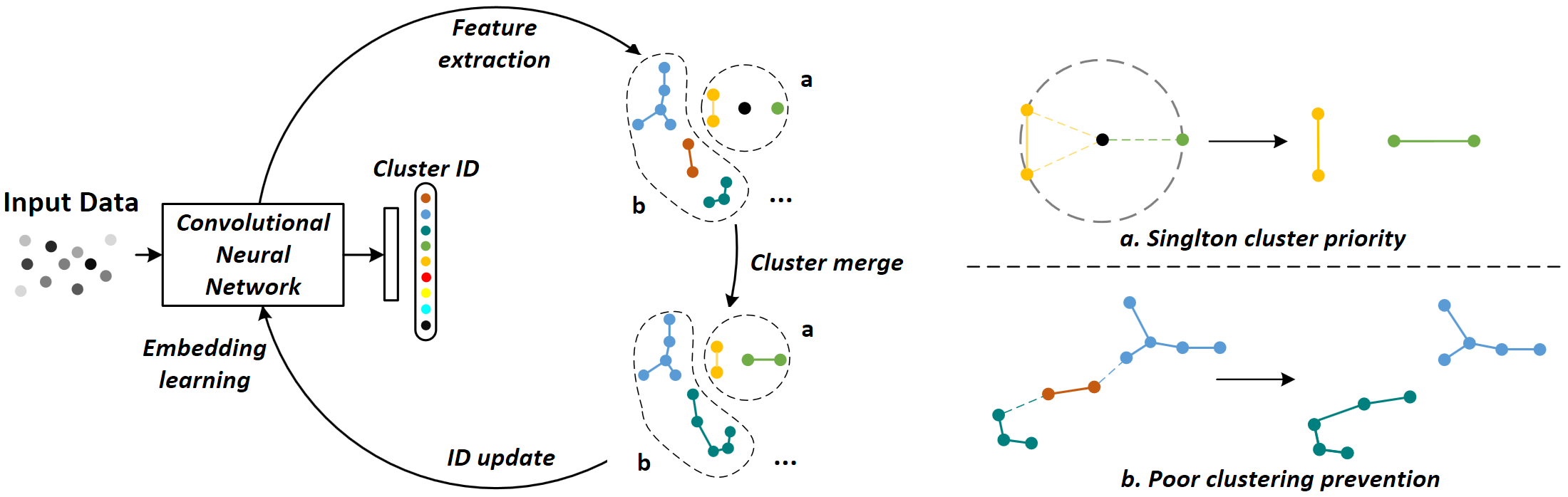}
    \caption{ The overall framework for the unsupervised learning. The left part shows the iterative process of our approach. The framework first extracts image features with a CNN model, then clustering is performed by the feature similarities and lastly image labels are updated with new cluster set to retrain the feature extractor. After the training is finished, the CNN extracts features on both query and gallery samples and then performs a distance based retrieval. The right part exhibits two properties of our dispersion criterion. \textbf{(a)} Singleton cluster priority. In this case, green sample is merged with black sample as it has less (zero) dispersion compared with that of orange cluster. \textbf{(b)} Poor clustering prevention. We consider a cluster poorly merged when it has high dispersion. The blue cluster is prevented from merging with brown because the resulting cluster would have a high dispersion. Instead the brown cluster is merged with the green one which together have less intra-dispersion. 
    (Best viewed in color)}
    \label{fig:framwork}
    \end{center}
\end{figure*}
\subsection{Unsupervised Domain Adaption}
In the absence of labeled data for a certain task, domain adaptation often provides an attractive option given that labeled data of similar nature but from a different domain is available. A main practice for adaptation is to align the feature distribution between the source and target domain \cite{tzeng2015simultaneous,long2015learning,sun2016deep,ganin2015unsupervised,sun2016return}. Likewise, some unsupervised cross-domain transfer learning methods \cite{peng2016unsupervised,zhong2019camstyle} have been studied in the field of person re-ID to deal with the misalignment between identities among different datasets. To better bridge this gap, Wang \etal\ \cite{wang2018transferable} first train with attributes on source domain and then learn a joint feature representation of both identity and attribute. A hetero-homogeneous learning approached is introduced in \cite{zhong2018generalizing} to align domain distributions. Another stream of work uses generative adversarial networks (GAN) to generate augmented images to reduce the dataset differences \cite{zhong2019camstyle,deng2018image}. Deng \etal\ \cite{deng2018image} explored image self-similarity and cross-domain dissimilarity for a target domain image translation. Differently, Zhong \etal\ \cite{zhong2019camstyle} exploited camera-to-camera alignment to perform image translation. These domain adaptation methods are all focused on the label estimation of target domain. One can see that the success of this category of approaches is based on an auxiliary labeled dataset. Compared to them, the unsupervised learning approach in this paper does not use any external data or annotations.

\subsection{Clustering Analysis} Clustering analysis is a long-standing approach to unsupervised machine learning. With the surge of deep learning techniques, recent studies have attempted to optimize clustering analysis and representation learning jointly for maximizing their complementary benefits \cite{caron2018deep,ghasedi2017deep,xie2016unsupervised,yang2017towards}. Fan \etal\ \cite{fan2018unsupervised} combines domain transfer and clustering for unsupervised re-ID task. They first train the model on an external labeled dataset which is used as a good model initialization. After that, unlabeled data samples are progressively selected for training according to their credibility defined as their distance to cluster centroids. However, this work relies on a strong assumption about the total number of identities. Aside from these methods that require auxiliary datasets or assumptions, Lin \etal\ \cite{lin2019bottom} proposed to apply a bottom-to-up framework for clustering, which  hierarchically combines clusters according to a predefined criterion. The merging in \cite{lin2019bottom} is based on a very simple minimum distance criterion with a cluster size regularization term. Different from their work, our dispersion criterion exploits feature affinities within and between clusters, which also has mutual interaction with the CNN model training process to reciprocate the model strength.

\section{Cluster Dispersion Criterion}
\label{sec:approach}
In this section, we start with some preliminaries  followed by our proposed criterion described in detail, and end with  discussions and comparisons with close work.

\subsection{Preliminaries}
Given an unlabeled training set $\mathcal{D} = \{x_i\}_{i=1}^N$ containing $N$ cropped person images, we aim to learn a feature embedding function $\phi(x_i;\theta)$ from $\mathcal{D}$ without any available annotations. The parameters $\theta$ are optimized iteratively using an objective function. This feature extractor can be later applied to the gallery set $\{x_i^g\}_{i=1}^{N_g}$ and the query set $\{x_i^q\}_{i=1}^{N_q}$ to obtain their feature representations for a distance based retrieval. The distance between each pair of images is defined as, $dist(x_i^q,x_i^g) = \|\phi(x_i^q;\theta) -\phi(x_i^g;\theta)\|$. For a higher distance based rank of a given pair, it is more likely that the pair belongs to the same identity. 

Supervised learning provides person identity labels $y_i$ for each input image $x_i$, and the feature embedding function is appended by a classifier $f(\phi;w)$ parameterized by $w$. Thus, $\phi(;\theta)$ can be learnt by optimizing the following objective function:
\begin{equation}
\label{eq:loss}
    \mathop {\min }\limits_{\theta,w} \sum_{i=1}^N l(f(\phi(x_i^q,\theta);w),y_i)
\end{equation}
where $l$ is the cross-entropy (CE) loss for classification. One shortcoming of CE loss is it does not explicitly minimizes  the intra-class distances. To this end, center loss is proposed that seeks to achieve within class compactness. 

Similar to center loss \cite{wen2016discriminative,ding2018feature}, repelled loss \cite{xiao2017joint,lin2019bottom} can act as a classifier $f$ which has the ability to jointly consider inter-class and intra-class variances by computing probability based on the feature similarity as follows:
\begin{equation}
    \label{eq:repel}
    p(y|x,V) = \frac{\exp(V_y^Tv/\tau)}{\Sigma_{j=1}^N \exp(V_j^Ty/\tau)},
\end{equation}
where $\tau$ is a temperature parameter that controls the softness of probability distribution over classes, $v$ is the $l_2$ normalized image feature obtained from $\phi(x;\theta)$, while $V$ is a lookup table (LUT) containing the centroid feature of each class. This LUT is updated on the fly by exponential moving average \cite{lucas1990exponentially} over training iterations, thus avoiding exhaustive feature extraction that is more computation intensive.
 
\subsection{Validity Guided Dispersion Criterion}

The main challenge towards using the above framework for an unsupervised setting lies in automatic label assignment for unlabeled data. Here, clustering comes as a natural choice as it aims to group similar entities together. In this paper, we propose a novel dispersion based agglomerative clustering approach based on a cluster validity consideration. The choice of affinity/dissimilarity measure between two clusters is the key to our proposed algorithm. In the task of person re-ID, which focuses on identifying images of the same identity, the inter and intra-cluster similarity should be considered for a reasonable merging. This requisite is fulfilled by a novel merging criterion used in our agglomerative clustering approach.

Given a cluster $\mathcal{C}$ scattered in feature space, we define its dispersion $d(\mathcal{C})$ as the average pairwise distance within:
\begin{equation}
    \label{eq:compactness}
    d(\mathcal{C}) = \frac{1}{n}\sum_{i,j\in \mathcal{C}} dist(\mathcal{C}_i,\mathcal{C}_j),
\end{equation}
where $n$ is the cardinality of cluster $\mathcal{C}$. As such, the dispersion between clusters can be written as follows:
\begin{equation}
    \label{eq:separation}
    d(\mathcal{C}_a,\mathcal{C}_b) = \frac{1}{n_a n_b}\sum_{i\in \mathcal{C}_a, j\in \mathcal{C}_b} dist(\mathcal{C}_{a_i},\mathcal{C}_{b_j})
\end{equation}
To jointly consider both intra- and inter-cluster dispersion, we have the dissimilarity between clusters $\mathcal{C}_a$ and $\mathcal{C}_b$ formulated as:
\begin{equation}
\label{eq:dissim}
D_{ab} = d_{ab}+ \lambda(d_a + d_b)
\end{equation}
where $d_{ab}$ and $d_a$ are used in place of $d(\mathcal{C}_a,\mathcal{C}_b)$ and $d(\mathcal{C}_a)$ for notation simplicity, and $\lambda$ is the trade-off parameter between two components. 

The former component $d_{ab}$ in Eq. \eqref{eq:dissim}, dispersion between clusters, is a measure for cluster dissimilarity. A cluster with low dispersion should be considered for merging as features from the same identity should be close in the feature space. The later component $d_a+d_b$, which is the sum of dispersion of both candidate clusters, serves as a regularizer to the former component. On one hand, it can help prioritize standalone data points for merging at the starting stages. On the other hand, this term can prevent escalating "poor" clustering as the high dispersion within-cluster can overbalance the inter-cluster term. In fact, this candidate cluster selection strategy controls the trade-off between the tendency to select spatially closer clusters ($\lambda \to 0$) or more compact clusters ($\lambda \to +\infty$).

\begin{table*}[htb]
\caption{Statistics of four datasets used in our experiment settings. "\# Samples" stands for the number of images for image-based datasets and number of tracklets for video-based datasets.}
\begin{center}
\begin{tabular}{l|c|cc|cc|cc}
\hline
\multirow{3}{*}{Dataset} &\multirow{3}{*}{Category} & \multicolumn{2}{c|}{Training} & \multicolumn{4}{c}{Testing} \\\cline{3-8}
 & & \multirow{2}{*}{\# ID} & \multirow{2}{*}{\# Samples} & \multicolumn{2}{c|}{Query} & \multicolumn{2}{c}{Gallery} \\\cline{5-8}
 & & &  & \# ID & \# Samples & \# ID & \# Samples \\\hline
Market-1501~\cite{zheng2015scalable} & Image-based &751 & 16,522 & 750 & 3,368 & 750 & 19,732 \\
DukeMTMC-reID~\cite{zheng2017unlabeled} & Image-based  & 702 & 16,522 & 702 & 2,228 & 1,110 & 17,661 \\
MARS~\cite{zheng2016mars} & Video-based & 625 & 8,298 & 626 & 1,980 & 636 & 12,180 \\
DukeMTMC-VideoReID~\cite{wu2019progressive} & Video-based & 702 & 2,196 & 702 & 702 & 801 & 2,636 \\
\hline
\end{tabular}
\label{tab:datasets}
\end{center}
\end{table*}
\begin{algorithm}[t]
   \caption{Dispersion based Clustering Algorithm}
   \label{alg:cud}
\begin{algorithmic}
   \STATE {\bfseries Input:} Training data $\mathcal{D} =\{x_i\}_{i=1}^N$, merging percentage $m \in (0,1)$, trade-off parameter $\lambda$, CNN model $\phi(\cdot,\theta_0)$
   \STATE {\bfseries Output:} Optimized model $\phi(\cdot ; \hat{\theta})$
   \STATE Initialize label $\mathcal{Y} = \{y_i=i\} _{i=1}^N$, Cluster number $C=N$, merge batch $k = N*m$
   \WHILE{$C > k$}
   \STATE Train model with $\{x_i\}$ and $\{y_i\}$ with Eq. \eqref{eq:loss}
   \STATE Calculate cluster dissimilarity matrix $\mathcal{P}(\mathcal{C})$
   \FOR{1:k}
   \STATE Select candidate clusters according to Eq. \eqref{eq:dissim} and merge them  
   \STATE Update matrix $\mathcal{P}(\mathcal{C})$ with Eq. \eqref{eq:update} and Eq. \eqref{eq:newintra}
   \STATE $C \leftarrow C-1$
   \ENDFOR
   \STATE Update $\mathcal{Y} $ with new cluster $\mathcal{C}$ using Eq. \eqref{eq:newlabel}
   \STATE Evaluate Performance $Perf$ on validation set.
   \IF{$Perf > Perf^*$}
   \STATE $Perf^* = Perf$
   \STATE Best model = $\phi(\cdot ; \hat{\theta})$
   \ENDIF
   \ENDWHILE
\end{algorithmic}
\end{algorithm}


\subsection{Matrix Update}
Proposed dispersion criterion can be calculated with a matrix updating algorithm. The input to this clustering process is the dissimilarity matrix $\mathcal{P}(\mathcal{C})$, also referred as the proximity matrix. It is a $C \times C$ matrix whose $(i,j)^{th}$ element equals the inter cluster dispersion $d(\mathcal{C}_i,\mathcal{C}_j)$ between $\mathcal{C}_i$ and $\mathcal{C}_j$. $\mathcal{P}(\mathcal{C})$ can be efficiently computed by first calculating an image pairwise distance matrix which is the outer product of stacked feature vectors obtained from deep networks and then aggregate them by cluster IDs.

At each clustering step, when two candidate clusters selected by Eq. \eqref{eq:dissim} are merged, the size of dissimilarity matrix $\mathcal{P}$ becomes $(C-1)\times(C-1)$. In one operation, two rows and columns of corresponding merged clusters $\mathcal{C}_a$ and $\mathcal{C}_b$ are deleted and a new row and a new column are added that contain the updated dissimilarity between the newly formed cluster $\mathcal{C}_q$ and an old cluster $\mathcal{C}_s$. The dissimilarity between $\mathcal{C}_q$ and $\mathcal{C}_s$ can be found using our dispersion definition, as follows:
\begin{equation}
    \label{eq:update}
    d_{qs} =\frac{n_a}{n_a+n_b} d_{as} + \frac{n_b}{n_a+n_b} d_{bs}.
\end{equation}
Correspondingly, the intra-cluster dispersion of the newly formed cluster $C_q$ is written as:
\begin{equation}
    \label{eq:newintra}
    d_{q} = \frac{n_a d_a + n_b d_b + n_a n_b d_{ab}}{n_a + n_b + n_a n_b}.
\end{equation}

\subsection{Learning Paradigm}
\label{subsec:update}
The overall learning paradigm with proposed criterion is presented as the left part in Fig. \ref{fig:framwork}, where two stages, \ie, embedding learning and clustering, take place on an alternating basis. This paradigm initiates with the {\em embedding learning stage} where each data point $x_i$ assigned a unique label $y_i$ and the CNN model is trained for a few epochs to learn the mapping. This choice of considering every single sample as an independent class is often referred as `sample specificity learning'. The key idea is that even with this naive supervision, neural networks still can automatically reveal the visual similarity correlation between different classes and yield a decent CNN initialization. 

In between embedding learning is the {\em clustering stage}. We consider a clustering stage to have $k$ {\em steps}, where top-$k$ cluster pairs with least dissimilarity defined by Eq. \eqref{eq:dissim} are to be merged. The $k$ is defined as the product $k=m*N$, controlled by a merging percent parameter $m$ (set of 0.05 percent of total number of samples in our experiments). After each step, the proximity matrix $\mathcal{P}(\mathcal{C})$ is updated with Eq. \eqref{eq:update} and Eq. \eqref{eq:newintra} introduced above. Before entering the next stage of CNN model training, samples in the entire training set are designated new labels according to their belonging to the resulting clusters as follows:
\begin{equation}
\label{eq:newlabel}
\mathcal{Y} = \{y_i = j,\quad  \forall x_i \in \mathcal{C}_j\}_{i=1}^N.
\end{equation}

The whole training procedure of our unsupervised person re-ID learning is summarized in Algorithm \ref{alg:cud}.

\subsection{Complexity Analysis} We analyze the per-stage complexity cost induced by the new criterion. For a clustering stage which performs $k$ merging operations, the computation complexity for image pairwise distance computation is $\mathcal{O}(N^2)$ and $\mathcal{O}(C^2)$ for cluster pairwise dissimilarity calculation. Afterwards, a cost of $\mathcal{O}(C\log C)$ is required for ranking and candidate selection. Lastly, a cost of $\mathcal{O}(kC)$ for $k$ step merging and proximity matrix update is incurred. So the total computation complexity for the whole stage comes together to $\mathcal{O}(N^2+C^2+C\log C+kC)$. Notice that since the cluster number $C$ decreases as the learning paradigm progresses, the computation complexity also decreases. Accordingly, the overall complexity is $\mathcal{O}(N^2)$, which means the main complexity comes from the inevitable sample-wise similarity calculation. All operations mentioned above can be computed by matrix manipulation on GPUs and one can use multiple GPUs for acceleration since our algorithm is suitable for parallelism.

\subsection{Discussion.} 

\textbf{The regularization term.}
The combination of the second component in Eq. \eqref{eq:dissim} brings two advantages to the clustering process stated as follows:

\textbf{{\em 1) Singleton cluster priority}}. Recall that each sample is assigned a unique label in the first round of CNN training in Sec \ref{subsec:update}, yielding all singleton clusters, \ie, clusters with only a single sample. However, the intrinsic matching and association property of the re-ID task requires that there exists at least two samples for a given identity. Therefore, singleton clusters cannot emerge during clustering by the problem definition and must be explicitly taken care of. Importantly, they should be dealt with at the first few  clustering stages as they may be further pushed away from points of their own identity as the CNN is trained to separate them. The priority shifting happens when two merging options have identical inter dispersion $d_{ab}$, consequently the standalone data point with less (zero) intra-cluster dispersion ($d_{a} = 0$ or $d_{b} = 0$) gets promoted. An illustration can be found in Fig.~\ref{fig:framwork}(a). 

\textbf{{\em 2) Poor clustering prevention}}. One disadvantage of the nesting property of agglomerative clustering is that there is no way to recover from a `poor' clustering that may have occurred in previous levels of the hierarchy \cite{gower1967comparison}. The addition of the regularization term helps to avoid this. Consider the case where a poor cluster formed in previous merging step, the high intra-cluster dispersion would prevent it from being selected for merging in the following turns, albeit it may have high rankings in intra-cluster dispersion based merging list. An illustration can be found in Fig.~\ref{fig:framwork}(b), brown cluster is merged with more distanced green cluster for their lower intra-dispersion.

\textbf{Comparison with close work.}
Our work shares a similar spirit as that of Bottom-up Clustering (BUC) \cite{lin2019bottom} and adopts an agglomerative clustering framework for the task of unsupervised person re-ID. We differ substantially in terms of cluster merge criterion. \textbf{\textit{On one hand}}, Lin \etal\ \cite{lin2019bottom} adopted minimum distance between cross cluster samples to measure their dissimilarity. It is known that the single linkage algorithm has a chaining effect, \ie, the dissimilarity $d_{qs}$ is obtained from $d_{as}$ and $d_{bs}$ whichever is smaller ($d_{qs} = \min\{d_{as},d_{bs}\}$). This implies it has a tendency to favor elongated clusters. Stretched clusters may hinder next iteration of model training with repelled loss which favours compact groups. \textbf{\textit{On the other hand}}, based on the presumption that training samples are evenly distributed among identities, Lin \etal\ \cite{lin2019bottom} proposed to use cluster cardinality as a diversity regularization term, however, this is also error-prone. As shown in Fig. \ref{fig:distribution}, the equal distribution of identity samples can hardly be assured in person re-ID. 
In contrast, our criterion works on the pairwise distance between individual data point which exploits the intra-cluster relations and bypasses the imbalanced data situation. Also, our criterion formulation can help in forming compact and well-separated clusters, which serves the same purpose as the repelled loss used in CNN model training. Two alternating stages pursuing the same goal of lower intra-cluster variation and higher inter-cluster separation form a reciprocal effect and speeds up the training process. The superiority of our proposed cluster dispersion criterion is evidenced through the numeric results provided in Sec. \ref{subsec:ablation}.



\begin{figure}[htb]
\centering
\subfigure[Market-1501]{%
\includegraphics[width=0.22\textwidth]{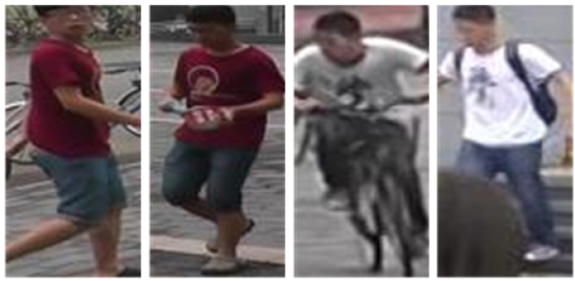}
\label{fig:market}}
\quad 
\subfigure[DukeMTMC-reID]{%
\includegraphics[width=0.22\textwidth]{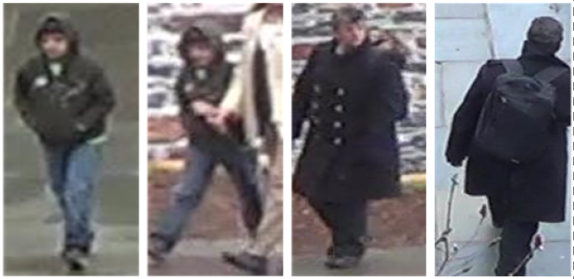}
\label{fig:duke}}

\subfigure[MARS]{%
\includegraphics[width=0.22\textwidth]{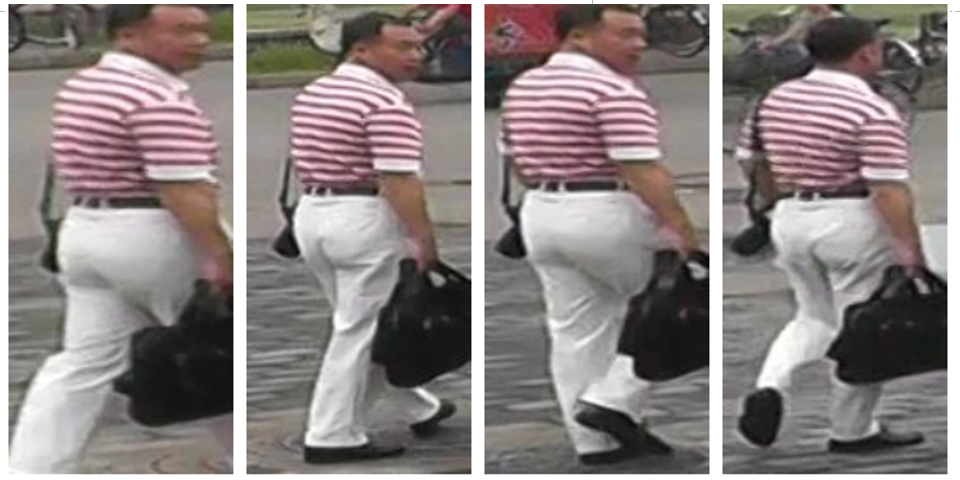}
\label{fig:mars}}
\quad 
\subfigure[DukeMTMC-VideoReID]{%
\includegraphics[width=0.22\textwidth]{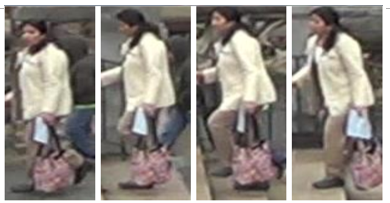}
\label{fig:dukevideo}}
\caption{Examples of person images from Market-1501, DukeMTMC-reID, MARS and DukeMTMC-VideoReID.}
\label{fig:datasets}
\end{figure}

\section{Experiments}
\label{sec:experiments}
In this section, we perform experiments on four large-scale person re-ID datasets to evaluate the effectiveness of our proposed approach.

\subsection{Datasets}
\label{subsec:datasets}
\textbf{Market-1501} \cite{zheng2015scalable} consists of 1,501 identities and 32,688 labeled images, among which 12936 images of 751 identities are used for training and 19,732 images of 750 identities are used for testing.

\textbf{DukeMTMC-reID} \cite{zheng2017unlabeled} contains 36,411 labeled images of 1,404 identities. Similar to Market1501, 702 identities are used for training and remaining identities as testing. Specifically, 16,522 training images, 2,228 query images and 17,661 gallery images.

\textbf{MARS} \cite{zheng2016mars} is a large-scale video-based dataset for person re-ID, which contains 17,503 video clips of 1,261 identities. The training set comprises of 625 identities while testing set comprises 636 identities.

\textbf{DukeMTMC-VideoReID} \cite{wu2019progressive} is a large-scale video-based dataset for person re-ID derived from DukeMTMC dataset. It has 2,196 tracklets of 702 identities for training, 2,636 tracklets of 702 identities for testing.

We list the statistics and some samples of these four datasets in Table \ref{tab:datasets} and Fig. \ref{fig:datasets}.

\subsection{Protocols}
\label{subsec:protocols}

\textbf{Training.}
To perform unsupervised learning on above mentioned person re-ID datasets, the training protocols are changed as described next. For image-based datasets, \ie, Market1501 and DukeMTMC-reID, training split remains the same except for the removal of identity labels. Similarly, for video-based datasets, \ie, MARS and DukeMTMC-VideoReID, the training samples are the tracklets of identities and each tracklet is treated as an individual. Note that no extra annotation information are used for model initialization or during our unsupervised feature learning.  
\begin{table*}[htb]
\caption{The effectiveness of our dispersion based criterion and comparison with minimum distance criterion.The  regularization terms are cluster size in BUC and intra-cluster dispersion in Ours.}
\begin{center}
\label{tab:component}
\begin{tabular}{l|cc|cc|cc|cc}
\hline
\multirow{2}{*}{Methods} & \multicolumn{2}{c|}{Market-1501} & \multicolumn{2}{c|}{DukeMTMC-reID} & \multicolumn{2}{c|}{MARS} & \multicolumn{2}{c}{DukeMTMC-VideoReID} \\ \cline{2-9}
 & rank-1 & mAP & rank-1 & mAP & rank-1 & mAP & rank-1 & mAP \\ \hline
BUC w/o regularization \cite{lin2019bottom} & 62.9 & 33.8 & 41.3 & 22.5 & 55.5 & 31.9 & 60.7 & 50.8 \\ 
BUC with regularization\cite{lin2019bottom} & 66.2 & 38.3 & 47.4 & 27.5 & 61.1 & 38.0 & 69.2 & 61.9 \\\hline
Ours w/o regularization & 66.2 & 38.7 & 48.2 & 27.5 & 59.8 & 37.2 & 71.8 & 63.2 \\ 
Ours with regularization & \textbf{69.2} & \textbf{41.3} & \textbf{51.5} & \textbf{30.0} &\textbf{64.3} & \textbf{43.8} &\textbf{75.2} & \textbf{66.1} \\ 
\hline
\end{tabular}
\end{center}
\end{table*}

\textbf{Testing.} 
When training is done, the CNN model with trained parameters is used as the feature extractor. The output activations from the penultimate layer of ResNet-50 are used as the person descriptor for image inputs, while the person descriptor for a tracklet input is the average of its frame features. These person descriptors are then used for a Euclidean distance based retrieval.

\textbf{Evaluation Metrics.} 
We evaluate our methods with rank-$k$ accuracy and mean average precision (mAP) on all four datasets. The rank-$k$ accuracy means that the correct match gets to be in the top-$k$ ranked list to count as `correct'. It represents the retrieval precision. The mAP value reflects the overall precision and recall rates. These two metrics provide a more comprehensive evaluation of the approach.

\begin{table*}[htb]
\caption{Comparison with the state-of-the-art methods on two image-based large-scale re-id datasets. The column "Labels" lists the supervision used by the corresponding method. "Transfer" means it uses an external dataset with annotations. "OneEx" denotes that only one image of each identity is labeled. "None" denotes no extra information is used.}
\begin{center}
\label{tab:market_duke}

\begin{tabular}{l|l|l|cccc|cccc}
\hline
\multirow{2}{*}{Methods} & \multirow{2}{*}{Venue} &\multirow{2}{*}{Labels}& \multicolumn{4}{c|}{Market-1501} & \multicolumn{4}{c}{DukeMTMC-reID} \\\cline{4-11}
 & & & rank-1 & rank-5 & rank-10 & mAP & rank-1 & rank-5 & rank-10 & mAP \\\hline
BOW\cite{zheng2015scalable} & ICCV'15 & \textbf{None} & 35.8 & 52.4 & 60.3 & 14.8 & 17.1 & 528.8 & 34.9 & 8.3 \\
OIM\cite{xiao2017joint}  & CVPR'17 & \textbf{None} & 38.0 & 58.0 & 66.3 & 14.0 & 24.5 & 38.8 & 46.0 & 11.3 \\
UMDL\cite{peng2016unsupervised} & CVPR'16 & Transfer & 34.5 & 52.6 & 59.6 & 12.4 & 18.5 & 31.4 & 37.6 & 7.3 \\
PUL\cite{fan2018unsupervised} & TOMM'18 & Transfer & 44.7 & 59.1 & 65.6 & 20.1 & 30.4 & 4604 & 50.7 & 16.4 \\
EUG\cite{wu2019progressive} & TIP'19 & OneEx & 55.8 & 72.3 & 83.5 & 26.2 & 48.8 & 63.4 & 68.4 & 28.5 \\
SPGAN\cite{deng2018image} & CVPR'18 & Transfer & 58.1 & 76.0 & 82.7 & 26.7 & 46.9 & 62.6 & 68.5 & 26.4 \\
TJ-AIDL\cite{wang2018transferable} & CVPR'18 & Transfer & 58.2 & - & - & 26.5 & 44.3 & - & - & 23.0 \\
BUC\cite{lin2019bottom} & AAAI'19 & \textbf{None} & 66.2 & 79.6 & 84.5 & 38.3 & 47.4 & 62.6 & 68.4 & 27.5 \\\hline
Ours & - & \textbf{None} & \textbf{69.2} & \textbf{83.0} & \textbf{87.8} & \textbf{41.3} & \textbf{51.5} & \textbf{64.6} & \textbf{70.1} & \textbf{30.0} \\
\hline
\end{tabular}
\end{center}
\end{table*}

\begin{table*}[htb]
\caption{Results on two video-based person re-identification datasets, MARS and DukeMTMC-VideoReID. The column "Labels" lists the supervision used by the corresponding method. "OneEx" denotes the each person in the dataset is annotated with one labeled example. "None" denotes no extra information is used.}
\label{tab:mars_duke}
\begin{center}
\begin{tabular}{l|l|l|cccc|cccc}
\hline
\multirow{2}{*}{Methods} & \multirow{2}{*}{Venue}& \multirow{2}{*}{Labels} & \multicolumn{4}{c|}{MARS} & \multicolumn{4}{c}{DukeMTMC-VideoReID} \\ \cline{4-11}
 & & & rank-1 & rank-5 & rank-10 & mAP & rank-1 & rank-5 & rank-10 & mAP \\ \hline
OIM\cite{xiao2017joint} & CVPR'17 & \textbf{None} & 33.7 & 48.1 & 54.8 & 13.5 & 51.1 & 70.5 & 76.2 & 43.8 \\ 
DGM+IDE\cite{ye2019dynamic}& ICCV'17 & OneEx & 36.8 & 54 & - & 16.8 & 42.3 & 57.9 & 69.3 & 33.6 \\ 
Stepwise\cite{liu2017stepwise}& ICCV'17 & OneEx & 41.2 & 55.5 & - & 19.6 & 56.2 & 70.3 & 79.2 & 46.7 \\ 
RACE\cite{ye2018robust} & ECCV'18 & OneEx & 43.2 & 57.1 & 62.1 & 24.5 & - & - & - & - \\ 
DAL\cite{chen2018deep} & BMVC'18 & OneEx & 49.3 & 65.9 & 72.2 & 23.0 & - & - & - & - \\ 
BUC\cite{lin2019bottom} & AAAI'19 &\textbf{None} & 61.1 & 75.1 & 80.0 & 38.0 & 69.2 & 81.1 & 85.8 & 61.9 \\
EUG\cite{wu2019progressive} & TIP'19 & OneEx & 62.8 & 75.2 & 80.4 & 42.6 & 72.9 & 84.3 & 88.3 & 63.3 \\  \hline
Ours & - &\textbf{None} & \textbf{64.3} & \textbf{79.2} & \textbf{85.1} & \textbf{43.8} & \textbf{75.2} & \textbf{87.0} & \textbf{90.2} & \textbf{66.1} \\
\hline
\end{tabular}
\end{center}
\end{table*}

\subsection{Implementation details}
In our experiments, we conventionally adopt ResNet-50 \cite{he2016deep} as the backbone architecture with pre-trained weights on ImageNet~\cite{deng2009imagenet}. A two layer fully connected layer is added on top of the penultimate layer (2048-d) of ResNet-50 for smaller feature embedding (1024-d). The last classification layer is replaced by the implementation of Eq. \eqref{eq:repel} with varying cluster numbers. All datasets share the exact same set of hyper-parameters if not specified. For CNN model training, we set the total training epoch is to be 20, batch size to be 16, dropout rate to be 0.5. The CNN model is optimized by Stochastic Gradient Descent (SGD) with momentum set to 0.9. Learning rate for parameters is initialized to 0.1 and decreased to 0.01 after 15 epochs. For clustering stages, the batch merging parameter $m$ is set to be 0.05 and the trade-off parameter $\lambda$ in Eq. \eqref{eq:dissim} is set to be 0.5. On image-based re-ID datasets, \ie, Market-1501 and DukeMTMC-reID, the whole training process takes less than 3 hours to finish with a Tesla V100 GPU. For video-based re-ID datasets, \ie, MARS and DukeMTMC-VideoReID, it takes about 4 hours for complete training.

\subsection{Ablation Study}
\label{subsec:ablation}
We evaluate the effectiveness of our proposed clustering criterion on all datasets and Table \ref{tab:component} summarizes the numerical results.

\textbf{The effectiveness of the inter-cluster dispersion term.} We evaluate the effectiveness of our inter-cluster dispersion term by comparing to a very close work BUC~\cite{lin2019bottom}. For fair comparison, we report results of BUC without its size regularization term in the first row of Table \ref{tab:component} and our proposed criterion without intra regularization is shown in the third row. It can be observed that, across all four datasets, our method outperforms BUC by a large margin of around 6\% in rank-1 accuracy and 7\% in mAP. This performance gain comes solely from the difference in cluster linkage criterion. Single linkage based minimum distance criterion essentially forms stretched and elongated clusters. While ours average pairwise distance has the capability to take into consideration wider context. This type of full linkage algorithm better discovers patterns underlying the dataset. Notably, our model without the second term manages to achieve comparable results to that of full BUC model (second row).

\textbf{The effectiveness of the intra-cluster dispersion term.} We further study the effects of the intra-cluster dispersion regularization term. Results of our full model can be found in the fourth row. The numerical increase indicates that the regularization term is helpful to further boost the performance. On Market-1501, the rank-1 accuracy is increased from 66.2\% to 69.2\% and mAP from 38.7\% to 41.3\%. Similar trend is observed across all datasets which advocates its effectiveness. The alliance of intra and inter dispersion in the full model can lead to a better feature representation learning.

\begin{figure}[htb]
\centering
\subfigure[Rank-1 accuracy]{%
\includegraphics[width=0.22\textwidth]{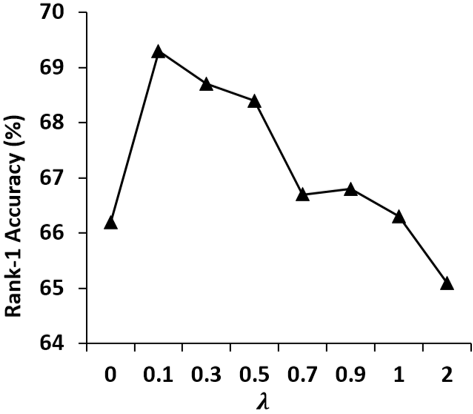}
\label{fig:rank1}}
\quad
\subfigure[mAP scores]{%
\includegraphics[width=0.22\textwidth]{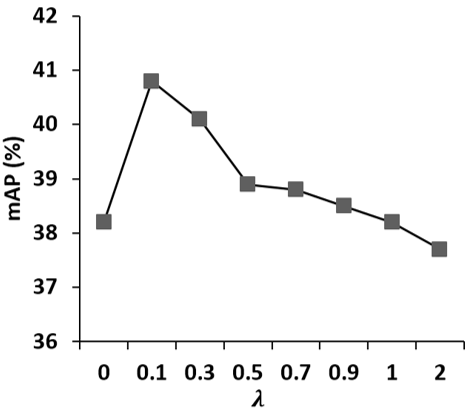}
\label{fig:map}}
\caption{Parameter study on Market-1501. We set varying values for trade-off parameter $\lambda$ and report Rank-1 accuracy and mAP changes in (a) and (b), respectively. }
\label{fig:lambda}
\end{figure}

\subsection{Algorithm Analysis}

\textbf{Analysis on the trade-off parameter.}
The regularization parameter $\lambda$ in Eq. \eqref{eq:dissim} balances  the importance of the intra-cluster and inter-cluster dispersion. We report results on Market-1501 dataset with varying $\lambda$ values in Fig.~\ref{fig:lambda}. It can be seen that the rank-1 accuracy first increases to its peak when $\lambda=0.1$ and then experiences a decline as shown in Fig.~\ref{fig:rank1}. A similar trend can be also found for mAP scores given in Fig.~\ref{fig:map}. It is plausible since this parameter can be interpreted as the preference in candidate cluster selection which emphasizes more on selecting clusters that are spatially close in feature space when $\lambda$ is relatively small, but more on selecting compact candidates as $\lambda$ increases. The best performance with $\lambda=0.1$ resonates with the empirical evidence that the inter dispersion should be the main clustering indicant coupled with reasonable regularization.



\begin{figure}[htb]
\begin{center}
    \centering
    \includegraphics[width=0.49\textwidth]{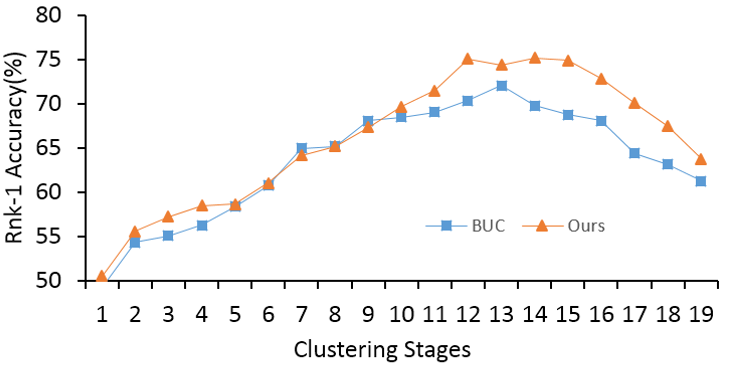}
    \caption{The rank-1 accuracy performance with respect to clustering stages on DukeMTMC-VideoReID. Our proposed criterion demonstrates a faster learning process and better robustness to cluster numbers, compared to BUC\cite{lin2019bottom}.}
    \label{fig:stages}
    \end{center}
\end{figure}
\textbf{Robustness.}
We further provide an analysis on the performance change over clustering iterations throughout the learning process to study its learning speed and robustness. Performance change with regards to clustering stages is shown in Fig. \ref{fig:stages}. One can see that we achieved best rank-1 accuracy at the $12^{th}$ clustering stage, one stage ahead of BUC. On DukeMTMC-VideoReID, one stage comprised of 100 pairs of merging; while on DukeMTMC-reID, one stage later convergence equates to 800$+$ more merging operations. This indicates that our algorithm has faster learning speed.

It can also be observed that the performances from both approaches  intertwined with each other before the $11^{th}$ clustering stage but diverged afterwards, with ours outperforming BUC by a relatively large margin. The tangled stages are the beginning stages where smaller clusters are being formed, progressively building up trustworthy identities and stimulating stable performance increase. After that, the algorithms started merging larger-sized clusters, during which inaccurate cluster merging could have a big influence on the performance. Those performance differences in the figure basically demonstrate the superiority of our approach. Another noticeable fact is that the performance of BUC dropped immediately after its peak while ours remained at the same level for few following stages before its decline. This phenomenon exhibited that our algorithm have better robustness to the resulting clustering numbers. Finally, the performance decline in the last few stages for both approaches is caused by the over-clustering of images from distinct identities since the resulting cluster number is far less than ground-truth identity number.

\begin{figure}[htb]
\begin{center}
    \centering
    \includegraphics[width=0.48\textwidth]{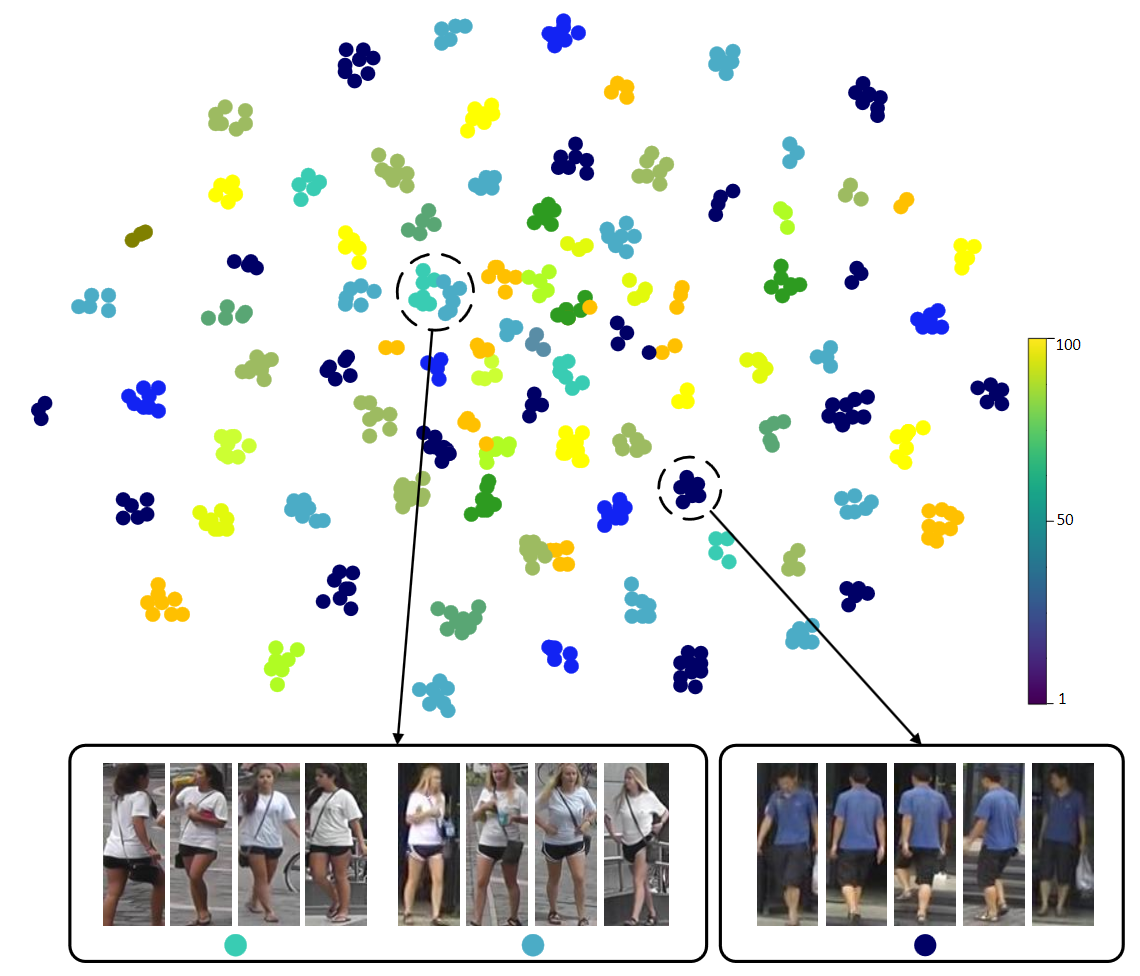}
    \caption{Qualitative illustration of clustering  on a reduced training set sampled from Market-1501 (100 identities, 1,657 images). T-SNE is adopted to visualize feature embeddings. Same color denotes same identity. The circled regions show two clustering cases \ie, negative (on left) and positive (on right) merging. We also show identity samples pointed by an arrow.  For the negative sample, it can be seen that two visually alike person are merged together.}
    \label{fig:tsne}
    \end{center}
\end{figure}

\textbf{Qualitative Analysis.}
We further provide a qualitative analysis to gain a better understanding of the clustering results by our proposed approach. In Fig \ref{fig:tsne}, T-SNE~\cite{maaten2008visualizing} is used to visualize the clustering results on a reduced dataset, which contains 1,657 images of 100 identities randomly sampled from Market-1501. It can be seen that in most cases, samples from the same identity are group together (see collections of same colored points). At the same time, there are some incorrect merging of identities. For example, see bottom left box in Fig \ref{fig:tsne} where two ladies with similar appearance (white t-shirts and dark sports wear) are clustered together. This highlights the fact that  visually alike identities are very difficult to disambiguate without any supervision.

\subsection{Comparison with state-of-the-art}
We evaluate our approach on both image-based and video-based person re-ID datasets and numerical results are reported in Table \ref{tab:market_duke} and Table \ref{tab:mars_duke}, respectively.

\textbf{Image-based Person re-ID.} Table \ref{tab:market_duke} summarizes the state-of-the-art unsupervised person re-ID results on Market-1501 and DukeMTMC-reID datasets. On Market-1501, we achieve the best performance among all listed approaches with \textbf{rank-1 = 69.2\%, mAP = 41.3\%}. Among the compared methods, OIM \cite{xiao2017joint} and BUC \cite{lin2019bottom} are evaluated under the fully unsupervised setting. It can be seen that our approach outperforms the state-of-the-art BUC by a margin of 3\%. Similar performance improvements can be observed on the DukeMTMC-reID dataset. 

Performance of some domain adaption and one-shot learning approaches are also reported, \eg\ TJ-AIDL \cite{wang2018transferable} and EUG \cite{wu2019progressive}. TJ-AIDL \cite{wang2018transferable} trains with attribute labels to learn a robust embedding encoding extra attribute information which is transferable, while EUG \cite{wu2019progressive} initializes model with one example labels and then progressively selects samples for training. In our experiment, we still surpass them by a relatively large margin (11\% and 13.4\% in rank-1 accuracy) even though external supervisions are provided in their settings. 

\textbf{Video-based Person re-ID.} The comparisons with state-of-the-art algorithms on video-based person re-ID datasets, MARS and DukeMTMC-VideoReID are reported in Table~\ref{tab:mars_duke}. On DukeMTMC-VideoReID, we achieved \textbf{rank-1=75.2\%,} \textbf{mAP=66.1\%}, exceeding our competitor BUC \cite{lin2019bottom} by 6\% and 4.2\% in rank-1 accuracy and mAP, respectively. This demonstrates a more stable generalization ability of our proposed clustering algorithm to different data distributions. We also managed to outperform all other competitive methods on MARS dataset with \textbf{rank-1=64.3\%, mAP=43.8\%}. These results illustrate the effectiveness of our proposed approach.

\section{Conclusion}
\label{sec:conclusion}
In this paper, we highlight the importance of quantifying cluster validity using robust statistical measures. We consider the problem of unsupervised person re-ID and propose a dispersion based criterion to assess the quality of automatically generated clusters. On one hand, the proposed criterion considers both intra- and inter-cluster dispersion and can achieve better clustering. The former dispersion term enforces compact clusters, while the latter ensures the separation between them. On the other hand, the proposed criterion can handle singleton clusters and prevent poor clustering. The overall extensive performance evaluations and ablation study illustrates the effectiveness of our proposed method.


\ifCLASSOPTIONcaptionsoff
  \newpage
\fi

{\small
\bibliographystyle{IEEEtran}
\bibliography{./bibtex/bib/IEEEabrv,./bibtex/bib/myrefs.bib}

\begin{thebibliography}{10}
\providecommand{\url}[1]{#1}
\csname url@samestyle\endcsname
\providecommand{\newblock}{\relax}
\providecommand{\bibinfo}[2]{#2}
\providecommand{\BIBentrySTDinterwordspacing}{\spaceskip=0pt\relax}
\providecommand{\BIBentryALTinterwordstretchfactor}{4}
\providecommand{\BIBentryALTinterwordspacing}{\spaceskip=\fontdimen2\font plus
\BIBentryALTinterwordstretchfactor\fontdimen3\font minus
  \fontdimen4\font\relax}
\providecommand{\BIBforeignlanguage}[2]{{%
\expandafter\ifx\csname l@#1\endcsname\relax
\typeout{** WARNING: IEEEtran.bst: No hyphenation pattern has been}%
\typeout{** loaded for the language `#1'. Using the pattern for}%
\typeout{** the default language instead.}%
\else
\language=\csname l@#1\endcsname
\fi
#2}}
\providecommand{\BIBdecl}{\relax}
\BIBdecl

\bibitem{yang2018person}
X.~Yang, M.~Wang, and D.~Tao, ``Person re-identification with metric learning
  using privileged information,'' \emph{{IEEE} Trans. Image Process.}, vol.~27,
  no.~2, pp. 791--805, 2018.

\bibitem{cho2018pamm}
Y.-J. Cho and K.-J. Yoon, ``Pamm: Pose-aware multi-shot matching for improving
  person re-identification,'' \emph{{IEEE} Trans. Image Process.}, vol.~27,
  no.~8, pp. 3739--3752, 2018.

\bibitem{feng2018learning}
Z.~Feng, J.~Lai, and X.~Xie, ``Learning view-specific deep networks for person
  re-identification,'' \emph{{IEEE} Trans. Image Process.}, vol.~27, no.~7, pp.
  3472--3483, 2018.

\bibitem{suh2018part}
Y.~Suh, J.~Wang, S.~Tang, T.~Mei, and K.~Mu~Lee, ``Part-aligned bilinear
  representations for person re-identification,'' in \emph{ECCV}, 2018.

\bibitem{sun2017svdnet}
Y.~Sun, L.~Zheng, W.~Deng, and S.~Wang, ``Svdnet for pedestrian retrieval,'' in
  \emph{ICCV}, 2017.

\bibitem{zhao2017spindle}
H.~Zhao, M.~Tian, S.~Sun, J.~Shao, J.~Yan, S.~Yi, X.~Wang, and X.~Tang,
  ``Spindle net: Person re-identification with human body region guided feature
  decomposition and fusion,'' in \emph{CVPR}, 2017.

\bibitem{zhou2017point}
S.~Zhou, J.~Wang, J.~Wang, Y.~Gong, and N.~Zheng, ``Point to set similarity
  based deep feature learning for person re-identification,'' in \emph{CVPR},
  2017.

\bibitem{farenzena2010person}
M.~Farenzena, L.~Bazzani, A.~Perina, V.~Murino, and M.~Cristani, ``Person
  re-identification by symmetry-driven accumulation of local features,'' in
  \emph{CVPR}, 2010.

\bibitem{liao2015person}
S.~Liao, Y.~Hu, X.~Zhu, and S.~Z. Li, ``Person re-identification by local
  maximal occurrence representation and metric learning,'' in \emph{CVPR},
  2015.

\bibitem{lisanti2015person}
G.~Lisanti, I.~Masi, A.~D. Bagdanov, and A.~Del~Bimbo, ``Person
  re-identification by iterative re-weighted sparse ranking,'' \emph{{IEEE}
  Trans. Pattern Anal. Mach. Intell.}, vol.~37, no.~8, pp. 1629--1642, 2015.

\bibitem{zhao2013unsupervised}
R.~Zhao, W.~Ouyang, and X.~Wang, ``Unsupervised salience learning for person
  re-identification,'' in \emph{CVPR}, 2013.

\bibitem{wang2014unsupervised}
H.~Wang, S.~Gong, and T.~Xiang, ``Unsupervised learning of generative topic
  saliency for person re-identification,'' in \emph{BMVC}, 2014.

\bibitem{kodirov2015dictionary}
E.~Kodirov, T.~Xiang, and S.~Gong, ``Dictionary learning with iterative
  laplacian regularisation for unsupervised person re-identification.'' in
  \emph{BMVC}, 2015.

\bibitem{fan2018unsupervised}
H.~Fan, L.~Zheng, C.~Yan, and Y.~Yang, ``Unsupervised person re-identification:
  Clustering and fine-tuning,'' \emph{ACM Trans. Multimedia Comput., Commun.
  Appl.}, vol.~14, no.~4, p.~83, 2018.

\bibitem{peng2016unsupervised}
P.~Peng, T.~Xiang, Y.~Wang, M.~Pontil, S.~Gong, T.~Huang, and Y.~Tian,
  ``Unsupervised cross-dataset transfer learning for person
  re-identification,'' in \emph{CVPR}, 2016.

\bibitem{deng2018image}
W.~Deng, L.~Zheng, Q.~Ye, G.~Kang, Y.~Yang, and J.~Jiao, ``Image-image domain
  adaptation with preserved self-similarity and domain-dissimilarity for person
  re-identification,'' in \emph{CVPR}, 2018.

\bibitem{lin2019bottom}
Y.~Lin, X.~Dong, L.~Zheng, Y.~Yan, and Y.~Yang, ``A bottom-up clustering
  approach to unsupervised person re-identification,'' in \emph{AAAI}, 2019.

\bibitem{khan2016unsupervised}
F.~M. Khan and F.~Bremond, ``Unsupervised data association for metric learning
  in the context of multi-shot person re-identification,'' in \emph{AVSS},
  2016.

\bibitem{kodirov2016person}
E.~Kodirov, T.~Xiang, Z.~Fu, and S.~Gong, ``Person re-identification by
  unsupervised $l_1$ graph learning,'' in \emph{ECCV}, 2016.

\bibitem{liu2017stepwise}
Z.~Liu, D.~Wang, and H.~Lu, ``Stepwise metric promotion for unsupervised video
  person re-identification,'' in \emph{ICCV}, 2017.

\bibitem{ma2017person}
X.~Ma, X.~Zhu, S.~Gong, X.~Xie, J.~Hu, K.-M. Lam, and Y.~Zhong, ``Person
  re-identification by unsupervised video matching,'' \emph{Pattern Recognit.},
  vol.~65, pp. 197--210, 2017.

\bibitem{wang2016towards}
H.~Wang, X.~Zhu, T.~Xiang, and S.~Gong, ``Towards unsupervised open-set person
  re-identification,'' in \emph{ICIP}, 2016.

\bibitem{ye2019dynamic}
M.~Ye, J.~Li, A.~J. Ma, L.~Zheng, and P.~C. Yuen, ``Dynamic graph co-matching
  for unsupervised video-based person re-identification,'' \emph{{IEEE} Trans.
  Image Process.}, 2019.

\bibitem{zhao2017person}
R.~Zhao, W.~Oyang, and X.~Wang, ``Person re-identification by saliency
  learning,'' \emph{{IEEE} Trans. Pattern Anal. Mach. Intell.}, vol.~39, no.~2,
  pp. 356--370, 2017.

\bibitem{yao2019deep}
H.~Yao, S.~Zhang, R.~Hong, Y.~Zhang, C.~Xu, and Q.~Tian, ``Deep representation
  learning with part loss for person re-identification,'' \emph{{IEEE} Trans.
  Image Process.}, 2019.

\bibitem{ma2012bicov}
B.~Ma, Y.~Su, and F.~Jurie, ``Bicov: a novel image representation for person
  re-identification and face verification,'' in \emph{BMVC}, 2012.

\bibitem{ma2012local}
------, ``Local descriptors encoded by fisher vectors for person
  re-identification,'' in \emph{ECCV}, 2012.

\bibitem{tzeng2015simultaneous}
E.~Tzeng, J.~Hoffman, T.~Darrell, and K.~Saenko, ``Simultaneous deep transfer
  across domains and tasks,'' in \emph{ICCV}, 2015.

\bibitem{long2015learning}
M.~Long, Y.~Cao, J.~Wang, and M.~I. Jordan, ``Learning transferable features
  with deep adaptation networks,'' in \emph{ICML}, 2015.

\bibitem{sun2016deep}
B.~Sun and K.~Saenko, ``Deep coral: Correlation alignment for deep domain
  adaptation,'' in \emph{ECCV}, 2016.

\bibitem{ganin2015unsupervised}
Y.~Ganin and V.~Lempitsky, ``Unsupervised domain adaptation by
  backpropagation,'' in \emph{ICML}, 2015.

\bibitem{sun2016return}
B.~Sun, J.~Feng, and K.~Saenko, ``Return of frustratingly easy domain
  adaptation,'' in \emph{AAAI}, 2016.

\bibitem{zhong2019camstyle}
Z.~Zhong, L.~Zheng, Z.~Zheng, S.~Li, and Y.~Yang, ``Camstyle: a novel data
  augmentation method for person re-identification,'' \emph{{IEEE} Trans. Image
  Process.}, vol.~28, no.~3, pp. 1176--1190, 2019.

\bibitem{wang2018transferable}
J.~Wang, X.~Zhu, S.~Gong, and W.~Li, ``Transferable joint attribute-identity
  deep learning for unsupervised person re-identification,'' in \emph{CVPR},
  2018.

\bibitem{zhong2018generalizing}
Z.~Zhong, L.~Zheng, S.~Li, and Y.~Yang, ``Generalizing a person retrieval model
  hetero-and homogeneously,'' in \emph{ECCV}, 2018.

\bibitem{caron2018deep}
M.~Caron, P.~Bojanowski, A.~Joulin, and M.~Douze, ``Deep clustering for
  unsupervised learning of visual features,'' in \emph{ECCV}, 2018.

\bibitem{ghasedi2017deep}
K.~Ghasedi~Dizaji, A.~Herandi, C.~Deng, W.~Cai, and H.~Huang, ``Deep clustering
  via joint convolutional autoencoder embedding and relative entropy
  minimization,'' in \emph{ICCV}, 2017.

\bibitem{xie2016unsupervised}
J.~Xie, R.~Girshick, and A.~Farhadi, ``Unsupervised deep embedding for
  clustering analysis,'' in \emph{ICML}, 2016.

\bibitem{yang2017towards}
B.~Yang, X.~Fu, N.~D. Sidiropoulos, and M.~Hong, ``Towards k-means-friendly
  spaces: Simultaneous deep learning and clustering,'' in \emph{ICML}, 2017.

\bibitem{wen2016discriminative}
Y.~Wen, K.~Zhang, Z.~Li, and Y.~Qiao, ``A discriminative feature learning
  approach for deep face recognition,'' in \emph{ECCV}, 2016.

\bibitem{ding2018feature}
G.~Ding, S.~Zhang, S.~Khan, Z.~Tang, J.~Zhang, and F.~Porikli, ``Feature
  affinity based pseudo labeling for semi-supervised person
  re-identification,'' \emph{{IEEE} Trans. Multimedia}, 2019.

\bibitem{xiao2017joint}
T.~Xiao, S.~Li, B.~Wang, L.~Lin, and X.~Wang, ``Joint detection and
  identification feature learning for person search,'' in \emph{CVPR}, 2017.

\bibitem{lucas1990exponentially}
J.~M. Lucas and M.~S. Saccucci, ``Exponentially weighted moving average control
  schemes: properties and enhancements,'' \emph{Technometrics}, vol.~32, no.~1,
  pp. 1--12, 1990.

\bibitem{zheng2015scalable}
L.~Zheng, L.~Shen, L.~Tian, S.~Wang, J.~Wang, and Q.~Tian, ``Scalable person
  re-identification: A benchmark,'' in \emph{ICCV}, 2015.

\bibitem{zheng2017unlabeled}
Z.~Zheng, L.~Zheng, and Y.~Yang, ``Unlabeled samples generated by gan improve
  the person re-identification baseline in vitro,'' in \emph{ICCV}, 2017.

\bibitem{zheng2016mars}
L.~Zheng, Z.~Bie, Y.~Sun, J.~Wang, C.~Su, S.~Wang, and Q.~Tian, ``Mars: A video
  benchmark for large-scale person re-identification,'' in \emph{ECCV}, 2016.

\bibitem{wu2019progressive}
Y.~Wu, Y.~Lin, X.~Dong, Y.~Yan, W.~Bian, and Y.~Yang, ``Progressive learning
  for person re-identification with one example,'' \emph{{IEEE} Trans. Image
  Process.}, 2019.

\bibitem{gower1967comparison}
J.~C. Gower, ``A comparison of some methods of cluster analysis,''
  \emph{Biometrics}, pp. 623--637, 1967.

\bibitem{ye2018robust}
M.~Ye, X.~Lan, and P.~C. Yuen, ``Robust anchor embedding for unsupervised video
  person re-identification in the wild,'' in \emph{ECCV}, 2018.

\bibitem{chen2018deep}
Y.~Chen, X.~Zhu, and S.~Gong, ``Deep association learning for unsupervised
  video person re-identification,'' in \emph{BMVC}, 2018.

\bibitem{he2016deep}
K.~He, X.~Zhang, S.~Ren, and J.~Sun, ``Deep residual learning for image
  recognition,'' in \emph{CVPR}, 2016.

\bibitem{deng2009imagenet}
J.~Deng, W.~Dong, R.~Socher, L.-J. Li, K.~Li, and F.-F. Li, ``Imagenet: A
  large-scale hierarchical image database,'' in \emph{CVPR}, 2009.

\bibitem{maaten2008visualizing}
L.~v.~d. Maaten and G.~Hinton, ``Visualizing data using t-sne,'' \emph{J. Mach.
  Learn. Res.}, vol.~9, no. Nov, pp. 2579--2605, 2008.

\end{thebibliography}
}
\end{document}